\def\year{2022}\relax
\documentclass[letterpaper]{article} % DO NOT CHANGE THIS
\usepackage{aaai22}  % DO NOT CHANGE THIS
\usepackage{times}  % DO NOT CHANGE THIS
\usepackage{helvet}  % DO NOT CHANGE THIS
\usepackage{courier}  % DO NOT CHANGE THIS
\usepackage[hyphens]{url}  % DO NOT CHANGE THIS
\usepackage{graphicx} % DO NOT CHANGE THIS
\usepackage{natbib}  % DO NOT CHANGE THIS AND DO NOT ADD ANY OPTIONS TO IT
\usepackage{caption} % DO NOT CHANGE THIS AND DO NOT ADD ANY OPTIONS TO IT
\usepackage{kotex}
\DeclareCaptionStyle{ruled}{labelfont=normalfont,labelsep=colon,strut=off} % DO NOT CHANGE THIS
\usepackage{algorithm}
\usepackage{algorithmic}
\usepackage{tabularx}
\usepackage{subfigure}

\usepackage[colorinlistoftodos, textwidth=2cm,textsize=tiny]{todonotes}

\usepackage{newfloat}
\usepackage{listings}
\floatstyle{ruled}
\newfloat{listing}{tb}{lst}{}
\floatname{listing}{Listing}
\title{Unraveling Code-Mixing Patterns in Migration Discourse:\\Automated Detection and Analysis of Online Conversations on Reddit}
\author {
    Fedor Vitiugin, Sunok Lee, Henna Paakki, Anastasiia Chizhikova, Nitin Sawhney
}
\affiliations{
    Department of Computer Science\\
    Aalto University, Finland\\
}

\usepackage{bibentry}
\begin{document}

\maketitle

\begin{abstract}
The surge in global migration patterns underscores the imperative of integrating migrants seamlessly into host communities, necessitating inclusive and trustworthy public services. Despite the Nordic countries' robust public sector infrastructure, recent immigrants often encounter barriers to accessing these services, exacerbating social disparities and eroding trust. Addressing digital inequalities and linguistic diversity is paramount in this endeavor. This paper explores the utilization of code-mixing, a communication strategy prevalent among multilingual speakers, in migration-related discourse on social media platforms such as Reddit.

We present Ensemble Learning for Multilingual Identification of Code-mixed Texts (ELMICT), a novel approach designed to automatically detect code-mixed messages in migration-related discussions. Leveraging ensemble learning techniques for combining multiple tokenizers' outputs and pre-trained language models, ELMICT demonstrates high performance (with F1 more than  0.95) in identifying code-mixing across various languages and contexts, particularly in cross-lingual zero-shot conditions (with avg. F1 more than 0.70). Moreover, the utilization of ELMICT helps to analyze the prevalence of code-mixing in migration-related threads compared to other thematic categories on Reddit, shedding light on the topics of concern to migrant communities.

Our findings reveal insights into the communicative strategies employed by migrants on social media platforms, offering implications for the development of inclusive digital public services and conversational systems. By addressing the research questions posed in this study, we contribute to the understanding of linguistic diversity in migration discourse and pave the way for more effective tools for building trust in multicultural societies.
\end{abstract}

\section{Introduction}

Between 2000 and 2020, global migration patterns witnessed significant shifts, with a 74\% growth, equivalent to approximately 37 million people. Europe experienced an increase of 30 million migrants, closely followed by North America with 18 million, and Africa with 10 million migrants~\citep{mcauliffe20221}. The escalating diversity emphasizes the importance of seamlessly integrating migrants into local processes and requires public services to facilitate smooth adaptation.

\begin{table}[t]
\scriptsize
\begin{tabularx}{\columnwidth}{p{0.15cm} p{1.5cm} p{5.5cm}}
 \hline
  & language & message \\
 \hline
 M1 &  English & Moved here from France because my wife is Finnish. I enjoy it enough to want to stay here. \\
 M2 &  code-mixed & WOW, sounds great to me, man. OK, kiitos paljon for your answer! \textbf{[EN]} \textit{WOW, sounds great to me, man. OK, thanks a lot for your answer!} \\
 M3 &  code-switched & Sun tarinas on yksi hyvä syy miksi oppia Englanti.  And of course... you know what happen next. \textbf{[EN]} \textit{Her/his story is one good reason to learn English. And of course... you know what happen next. }.\\
 M4 &  Finnish & Moikka! En oo sujuva suomen kielessä ja juuri nyt opiskelen suomen lukiossa. \textbf{[EN]} \textit{Hello! I'm not fluent in Finnish, and right now, I'm studying at a Finnish high school.} \\ 
 \hline
\end{tabularx}
\caption{Examples of code-mixed and non-mixed messages for English-Finnish pair. (\textit{Note: messages were paraphrased for anonymity.})}
\label{table:examples}
\end{table}

The Nordic countries have well-functioning and mostly equitable public sector services, earning the trust of a majority of citizens. However, this trust and efficiency do not always extend to recent immigrants and migrant communities residing in the region. For many migrants, these public services may seem inaccessible, lacking inclusivity or trustworthiness, which significantly undermines their integration~\citep{yeasmin2020immigration, intke2015migrant}.

Digital inequalities among specific groups can exacerbate social disparities, further marginalizing them and potentially undermining trust~\citep{madianou2015digital}. Therefore, it's vital for local municipalities to prioritize objectives like enhancing integration, promoting inclusion, and supporting migrant communities. Many cities are actively working to create innovative digital public services, such as chatbots, especially in areas like healthcare, employment, and social services. However, it's essential to ensure that these services are accessible to users with diverse linguistic backgrounds and varying levels of digital literacy. 

In the public sector and non-profit organizations assisting migrants, language is not restricted to a single mode of expression. Multilingual speakers tend to interleave two or more languages when communicating, a phenomenon known as code-mixing. This strategy has become increasingly prevalent in today's diverse linguistic and cultural landscape~\cite{gumperz1982discourse}. Due to this communication style, migrants naturally lean towards code-mixing to more effectively convey their circumstances and context.

A recent study highlights the complex linguistic practices employed by migrants in computer-mediated communication~\cite{mcentee2023multilingual, ikeh2023social}. Social media platforms provide bilingual users with a dynamic space to navigate their multiple identities online post-resettlement~\cite{harwood2019understanding, gardner2020contact}. Our research focuses on the Reddit platform, selected not only for the availability of data collection but also for its community-based structure and user-generated thread labels, simplifying content analysis.

Table~\ref{table:examples} demonstrates examples of various code-mixed, code-switched, and non-mixed messages. 
M1 and M4 are prototypical single-language messages in English and Finnish, respectively. 
M2 is an example of a code-mixed message, where the user included the Finnish phrase ``kiitos paljon'' instead of ``thanks a lot'' in his English message.
Finally, M3 is an example of a code-switched message, where the user starts their text with a Finnish sentence to explain the situation and continues with an English sentence. We will explain the difference between code-mixing and code-switching in the next section.

The emergence of Multilingual Large Language Models (LLMs) has demonstrated exceptional capabilities across various tasks~\cite{chang2023survey, zahera2023using}, showcasing state-of-the-art performance through zero-shot or few-shot methods. While extensive research has explored their monolingual capabilities, their potential in cross-lingual communication remains relatively unexplored~\citep{zhang2023multilingual}. However, current intelligence-based conversational systems often fail to meet the communicative expectations of multilingual migrant users, resulting in linguistic and cultural barriers. Consequently, there is an urgent need for the public sector to evolve these systems, considering the communication needs of migrants.

We explore migrants’ information requests shared on social media. Our paper addresses the following research questions:

\textbf{RQ1:} Can we automatically identify code-mixed social media messages from Reddit related to migration to Finland?

\textbf{RQ2:} How proficiently can the proposed approach identify instances of code-mixing in social media conversations in cross-lingual zero-shot conditions

\textbf{RQ3:} Which content topics exhibit a high proportion of code-mixed messages, and what are the differences in code-mixing usage between migration-related threads and threads in other user-defined categories?

To tackle the first two questions, we introduced a flexible approach named \textit{Ensemble Learning for Multilingual Identification of Code-mixed Texts (ELMICT)}, which relies on ensemble learning techniques ~\cite{abimannan2023ensemble}. This method effectively detects code-mixed social media messages. Our model integrates outputs from multiple tokenizers and fine-tuned pre-trained language models to identify texts containing code-mixing. We illustrate that using tokenizers or fine-tuned models separately yields lower performance and is less robust, particularly for texts containing out-of-vocabulary tokens. In addressing the third question, we calculate the proportion of code-mixing messages across various topics, including migration, tourism, politics, and general discussions.

The subsequent section of this paper will first present related research, followed by an explanation of our proposed method for detecting code-mixed social media messages and the setup for topic modeling of detected messages. Following this, we will detail our experimental setup and analyze the results. Finally, we will offer our conclusions and outline potential future work.

\section{Related Work}

In this section, we will discuss relevant works about code-mixing, and research on its usage in migrant communication. We also discuss methods for code-mixed text identification and the application of these methods for different tasks.

\subsection{Code-Mixing and Code-Switching}

Central to this work is the linguistic concept called ``code-mixing'', how it differs from ``code-switching'' and other language alternating techniques. Both are commonly used throughout the world, and are especially crucial for communities of migrants, expats, bilinguals, etc~\cite{gardner2020contact, mcentee2023multilingual}. These occur when two languages are used spontaneously in one sentence or expression.

Although the main purpose of our work is to research code-mixing in migration-related communication, we also wish to provide key definitions and discuss the differences between code-mixing and code-switching, as these are crucial for this study. More detailed information on these phenomena provided in related linguistics research cited in this subsection. 

Many scholars have attempted to define code-switching and code-mixing.
Weinreich, a leading researcher on bilingualism, has claimed that ``the ideal bilingual is someone who is able to switch between languages when required to do so by changes in the situation but does not switch when the speech situation is unchanged and \textit{certainly not within a single sentence}''~\cite{weinreich1953language}.
Specialists in code switching, however, recognize code switching as a functional practice and as a sign of bilingual competence~\cite{toribio2001emergence}.
Competence includes two aspects: fluency in speaking two or more languages and comprehensive understanding of them, even if speaking fluently is not necessary. It's evident that code-switching requires a high level of proficiency in multiple languages, rather than being a consequence of insufficient knowledge in one or the other language~\cite{poplack2000english}.

Code-switching refers to the ``use of two or more languages in the same conversation, usually within the same conversational turn, or even within the same sentence of that turn''~\cite{myers1993social}. Code-switching is the shifting by a speaker from language A to language B.

There are varying definitions of code-mixing. It's described as instances where a mix between the grammar of one language and another language is employed without changing the grammar of the initial language used~\cite{mabule2015code}.
On the other hand, ``Conversational code-mixing involves the deliberate mixing of two languages without an associated topic change''~\cite{wardhaugh2021introduction}. The definition indicates that code-mixing is typically used as a solidarity marker in multilingual communities. Similarly, according to other views, in code-mixing speakers switch between languages even within words (e.g. Spanglish or Finglish as a mixture of the English and Spanish or English and Finnish languages relatively) and/or phrases~\cite{milroy1995one, auer2013code}

In this paper, the term ``code-mixing'' is used to indicate \textit{a switch between languages, in which a single word or phrase from one language (here: Finnish, Spanish, or Korean) is integrated into another language (here: English)}.

\subsection{The Role of Code-Mixing in Migrant Communication}

Code-mixing among multilingual speakers commonly observed in close relationships, particularly when speaking with friends and family who share similar linguistic and cultural backgrounds~\cite{wulandari2021code}. However, speakers tend to avoid code-mixing if they're unsure how their interlocutors will react. Moreover, even when speakers are aware of their conversational partner's language proficiency, they may adjust their language usage to match the partner's code-mixing style and frequency, especially if trust is perceived to be lacking~\cite{kusumawati2023conversation}. This highlights how code-mixing serves as an indicator of trust and intimacy levels among multilingual speakers~\cite{choi2023toward}.

From the civil service practitioners' side, it is critical to make sure that services are accessible at the user experience level and linguistically, rather than broader aspects of its design and impact. Practitioners cited the lack of staff diversity and linguistic exclusion as the main challenges for better inclusion of citizens in such services~\cite{drobotowicz2023practitioners}. On the other hand, migrants may encounter challenges, particularly in critical contexts such as local government offices and hospitals, which place greater demands on language proficiency~\cite{st2023social}. Moreover, personification of the conversational agent could increase engagement~\cite{ostrowski2021small}, increase trust and relationships~\cite{schaefer2016meta, luria2019re}.

Conversational agents fail to understand users for many reasons, multilingual users often blame their unique speech behavior—code-mixing and drop the conversation or think they have lost control of the device because they do not understand the reason for the failure~\cite{ponnusamy2022feedback}. Experiences like this could greatly diminish the users’ well-established trust and intimacy with the conversational agent. For this reason, a code-mixing conversational agent should be designed to make clear statements and detailed explanations of their failure to prevent the multilingual users from getting frustrated by unnecessary misunderstanding~\cite{yap2021hci}.

Recent study participants prefer their agent to avoid unnecessary code-mixing but understand its usage in certain contexts. This preference originates from experiences where they were perceived as code-mixing due to language limitations and the importance of trust for acceptance in relationships. Additionally, designers could enhance trust and intimacy with code-mixing users by giving the agent a persona with diverse cultural or language backgrounds and similar code-mixing skills. This would enable users to contact the agent, similar to how they interact with other multilingual individuals~\cite{choi2023toward}.

\subsection{Code-Mixed Data Processing}

Recently, there has been a growing interest in the development of language models and technologies tailored for handling code-mixed content. Researchers have delved into exploring joint models capable of simultaneously performing language identification and part-of-speech tagging~\cite{barman2016part}. This dual-level language identification spans both word and sentence levels~\cite{rani2022mhe}. A method, based on the UDLDI model, employs a CNN architecture that incorporates enriched sentence and word embeddings~\cite{goswami2020unsupervised}.

Addressing the complexities of code-mixed content, certain studies have simplified texts by transforming them into a monolingual form through back-transliteration~\cite{dowlagar2021pre, gautam2021comet}. However, the efficacy of these techniques heavily relies on the accuracy of transliteration and translation methods employed.

Transfer Learning approaches have gained widespread attention in leveraging pre-trained language models for analyzing code-mixed data~\cite{aguilar2019english, krishnan2021multilingual}. Yet, the substitution of tokens in cross-lingual transfer learning can introduce grammatical inconsistencies in the resultant sentences, potentially impairing performance on token-sensitive tasks. To overcome this challenge, token-alignment techniques have emerged, facilitating not only token replacement but also considering contextual similarity to ensure grammatical coherence in both training and inference stages~\cite{feng2022toward}. The word segmentation method has shown promising results in code-mixed data processing. Utilization of a linguistics-based toolkit is maintaining the quality of monolingual translation with Hokkien-Mandarin code-mixed texts, widespread among Chinese immigrants~\cite{lu2023exploring}.

A review of recent literature underscores a pronounced emphasis on the tokenization issue. Indeed, accurate tokenization and word segmentation significantly enhance performance in code-mixing-related tasks. Furthermore, many studies have used synthetic training data, posing challenges for further analysis of real-world scenarios where users employ code-mixing in their communication.

\section{Method}

This study aims to explore the usage of code-mixing by migrants in social media. In this section, we present a supervised-learning classification model for detecting code-mixing in social media and describe methods used for analyzing code-mixed texts.

\subsection{Text Classification}

\begin{figure}[t]
\centering
\includegraphics[width=\columnwidth]{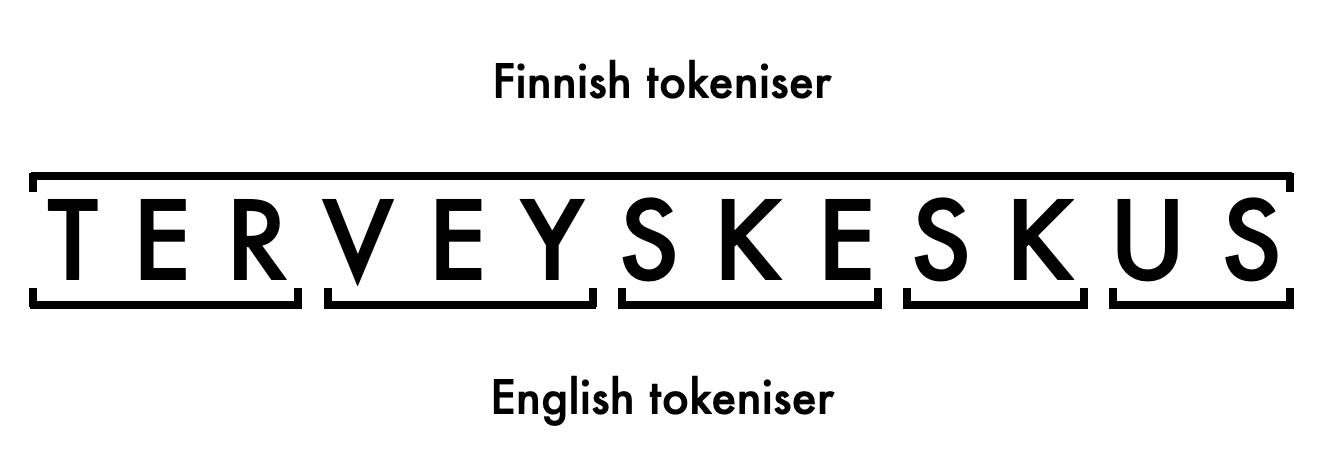}
\caption{Example of differences in single-language pre-trained model tokenizer outputs.} 
\label{fig:tokenization}
\end{figure}

Recent work in text classification analysis clearly demonstrates the necessity of precise word segmentation and tokenization. Compound words, which are quite rare in the majority of languages, play a significant role in Finnish. Figure~\ref{fig:tokenization} demonstrates the difference in English and Finnish pre-trained tokenizer outputs for the word ``terveyskeskus'' which means ``public health center''. This word is widely used not only by locals but also migrants, and plays an important role in their daily vocabulary.

The Ensemble Learning for Multilingual Identification of Code-mixed Texts (ELMICT) model aims to merge pre-trained language models with features generated by tokenizers through ensemble modeling. To ensure the classification model receives comprehensive information, we experimented with various combinations of tokenizer outputs, ultimately retaining four of them:
\begin{itemize}
    \item English BPE-tokenizer -- English language is used as the basic language in our datasets, so we choose the tokenizer from the most popular~\footnote{based on HuggingFace.com model popularity statistics} English transformer model;
    \item local language BPE-tokenizer -- Finnish, Korean, or Spanish BPE-tokenizer for related datasets;
    \item multilingual BPE-tokenizer -- we found that for some cases, multilingual tokenizers are also providing correct outputs and include multilingual BERT in our model;
    \item whitespace tokenizer -- NLTK whitespace tokenizer provides additional information as the most naive method.
\end{itemize}

Two other components of the proposed model are a language detection tool and a fine-tuned pre-trained transformer model for code-mixing detection. For language detection, lingua Python library~\footnote{https://github.com/pemistahl/lingua-py} was used in mixed-language mode. The model received information about the existence of English and local words\/phrases in the target text. XLM-RoBERTa was used for contextual detection of code-mixing. We fine-tuned a pre-trained model for the sequence classification task on English-Finnish texts (for both monolingual and cross-lingual tasks).

\begin{figure}[t]
\centering
\includegraphics[width=\columnwidth]{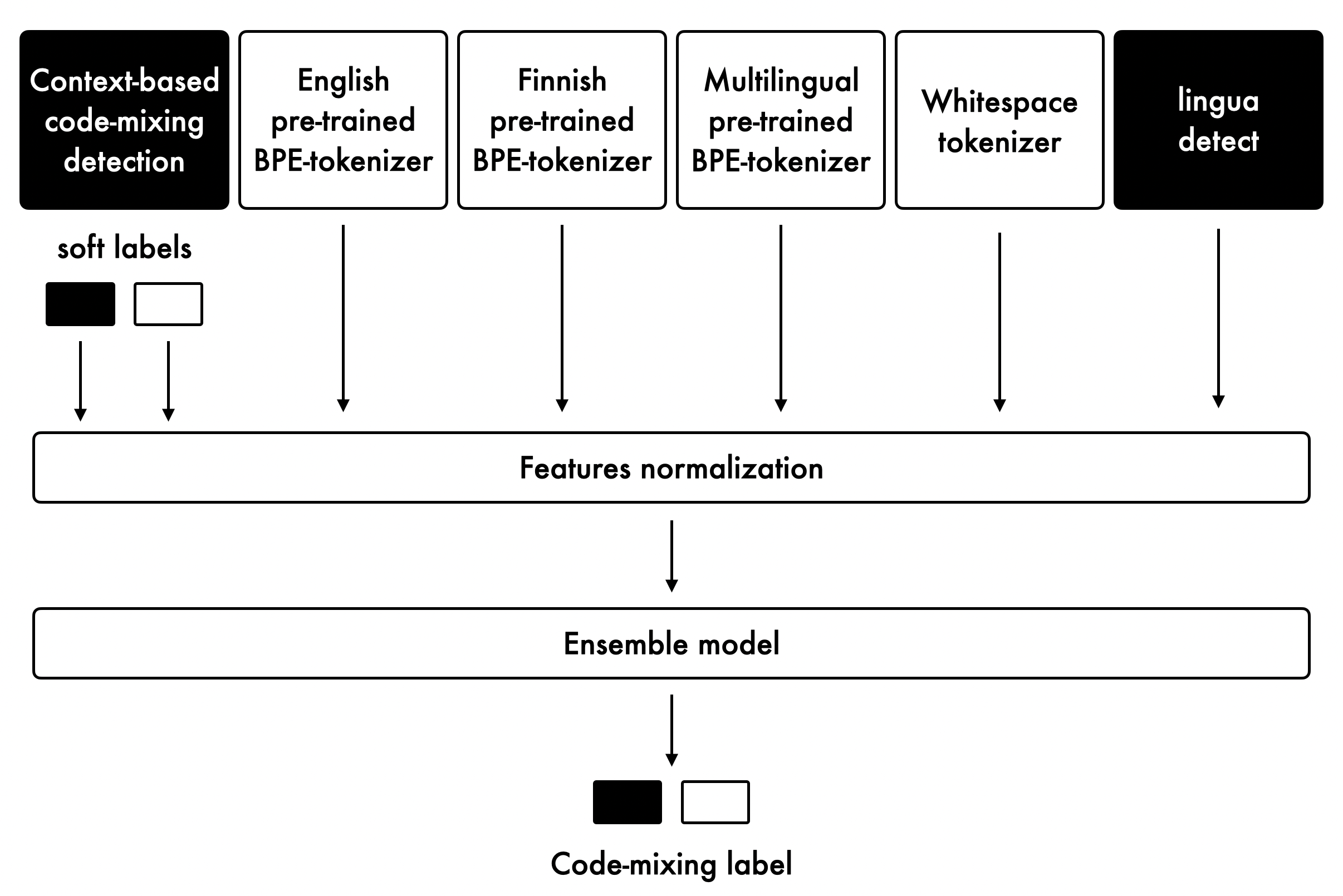}
\caption{ELMICT model architecture.} 
\label{fig:architecture}
\end{figure}

The architecture of ELMICT model presented in Figure~\ref{fig:architecture} has combined two approaches: contextual and feature-based. For the contextual approach, we fine-tune the multilingual pre-trained large language model. Soft labels output from the fine-tuned model were used as features for the ensemble model. As features, we used information extracted by 4 tokenizers. 

Our approach is based on the intuition that specialized tokenizers (Finnish tokenizer for a word in Finnish) will split relevant text into tokens more accurately, while unspecialized tokenizers (like English applied for word in Finnish) will generate more tokens (parts of word). It means that when an ensemble learning model receives the result of tokenization from different models, it can track which tokenization split of out-of-vocabulary tokens was done wrong.

\subsection{Topic Modeling}

To analyze the difference in migrant-related and general Reddit posts, we applied BERTopic technique~\cite{grootendorst2022bertopic} for topic modeling. The method utilizes BERT embeddings~\cite{devlin2018bert} to cluster the texts and leverages c-TF-IDF algorithm to further generate topic representations. 

First, we utilized sentence embeddings to convert input documents into numerical representations, enabling the capture of semantic similarity between documents. Second, we employed the UMAP dimensionality reduction algorithm~\cite{mcinnes2018umap} to address the high dimensionality of embeddings, which can make clustering challenging due to the curse of dimensionality. The clustering algorithm was used HDBSCAN~\cite{mcinnes2017hdbscan}, the default for BERTopic. Third, we experimented with various topic representation parameters and decided to use uni- and bi-grams only. Finally, we tested different minimal topic sizes and determined a threshold of 0.3\% of the original dataset size. These steps resulted in not only a high coherence score of 0.8 but also in topics that are interpretable by humans, which we further analyze in-depth.

%We ran the algorithm on the English-Finnish code-mixing dataset and calculated proportion of code-mixing for each topic and content \textit{flair} -- user-assigned labels for Reddit threads.

\subsection{Model Implementation}

We maintain the same number of layers as the original pre-trained model -- 24 layers for XML-RoBERTa~\cite{conneau2019unsupervised}. 
For the model's fine-tuning, we used \(0.5*10^{-5}\) learning rate, 10 epochs. 
The number of frozen layers for each model was detected by grid search.
The model was trained on NVIDIA A100-SXM4 with 40Gb GPU RAM.

\section{Experiment Setup}

\subsection{Data collection and annotation}

We collected posts and comments through the official Reddit API~\footnote{https://www.reddit.com/dev/api/} from three country-related communities (subreddits): \textit{r/Finland}, \textit{r/korea}, and \textit{r/GoingToSpain}. All three communities primarily use English, making them more accessible for migrants.  Each community has user-generated topic-related labels known as ``flair'', including migration-related flair.

All messages collected from location related communities were manually annotated by one human assessor with living experience in the corresponding area and language proficiency both in English and the code-mixed language. Additionally, we enlisted two individuals residing in each location to label 100 random messages from their respective communities to calculate assessor agreement. The Krippendorff's alpha for \textit{r/GoingToSpain} is 0.87, for \textit{r/Finland} is 0.75, and for \textit{r/korea} -- 0.92.

The labeling task was to assign one of the two classes for determining whether a given message is code-mixed or not for a target language pair. 
For uncertain words, the authors consulted with individuals, who are both proficient or native in target languages and currently living in the target country.

Before dataset annotation, we conducted a simple preprocessing step to filter out all uninformative tweets (based on manual analysis of a random sample,  more than 92\% of messages with length $\leq$ 4 tokens are uninformative).
Table~\ref{table:dataset} presents the quantity of train and test instances for each category, as well as unlabeled text entities.

\begin{table}
\scriptsize
\centering
\begin{tabular}{| p{1.5cm} | p{1cm} | p{1cm} | p{1.2cm} | p{1.5cm} |}
\hline
Community & Unlabelled &  Migration & Non-mixed & Code-mixed \\
\hline
\textit{r/GoingToSpain} & 39871 & 3514 & 878 & 122 \\
\textit{r/Finland}  & 174212 & 12632 & 2055 & 249 \\
\textit{r/korea} & 130388 & 3337 & 973 & 27 \\
\hline
\end{tabular}
\caption{Dataset statistics.}
\label{table:dataset}
\end{table}

During the process of labelling data related to Finland, several types of Finnish concepts were detected. The first group consists of cultural concepts and includes words like ``sisu'' (strength of will), ``handknit villasukat'' (hand knitted wool socks as marker of coziness), and ``mummola'' (grandmother's house). The second group contains words related to civil organizations and public services, like ``tilastokeskus'' (national statistical institution in Finland), ``terveyskeskus'' (public health center). The third group contains figurative compound words: ``piruntorjuntabunkkeri'' (church), ``betonihelvetti'' (concrete buildings). The final group is obscene language and slang: ``mamu''; ``ryssä''. Code-mixing generally occurs without special marking within the sentence, or sometimes marked by quotation marks.

In the context of texts related to migration in Korea, code-mixing has been observed predominantly with the use of Korean terms that reflect specific cultural and social contexts: ``mukbanger'', ``닭발'', ``hagwon'', and ``chaebol'', etc. In code-switching texts, Korean words are romanized, meaning they are transcribed phonetically into English, rather than being written directly in Hangul, the Korean alphabet. For example, due to Korean culture's unique practice of using specific titles instead of names to address someone, terms like ``unnie'', which means older sister, or referring to a child's father by combining ``Papa'' with the child's name, are used. Additionally, ``mukbanger'', which refers to a YouTuber who broadcasts their eating, and ``닭발'' (translated as ``chicken feet''), a word included to represent a facet of Korean food culture, illustrate the expression of cultural phenomena related to food that originated in Korea. Similarly, although ``hagwon'', denoting a private tutoring academy, can be translated into English, its use more precisely reflects Korea's unique educational culture. Moreover, ``다문화'' (damunhwa) is used to refer to people from diverse cultural backgrounds within Korean society; although 'migrant' exists as an English equivalent, ``damunhwa'' is used to convey the societal context more accurately. Notably, terms like ``chaebol'' (representing rich people or conglomerates) and ``JY Lee'', a quintessential figure in Korean chaebol culture, are utilized to denote Korea's distinctive corporate culture.

In the Spanish migration context, the majority of cases in which there was code-switching occurred are specific bureaucratic terms like ``extranjeria'' (foreigner), ``empadronamiento'' (census), ``pareja de hecho'' (domestic partnership) etc. These words and phrases do not have a direct English translation and in the context of conversations related to migration, it is important to be precise with the terms, so the users use the right Spanish terminology. Interestingly, sometimes they do that with terms that could be easily translated to English: ``Generally you should be fine with the seguridad social (social security)...''
Other, much less frequent cases include the insertion of Spanish slang ``guirris'' (tourists from Northern Europe or UK) and the usage of greetings (Hola (Hi)! at the beginning of a message in English).

\subsection{Schemes}

To evaluate the proposed method, we compared it with several state-of-the-art models.
The full list of proposed modeling schemes for evaluation is the following (* denotes our proposed models and others are the baselines):

\begin{itemize}
    \item lingua -- library for language identification based on model and data provided by ~\cite{biemann2004language};
    \item Random Forest -- classification model outputs of ensemble of tokenizers;
    \item Adaptive Boosting -- classification model outputs of ensemble of tokenizers;
    \item Gradient Boosting -- classification model outputs of ensemble of tokenizers;
    \item XLM-RoBERTa -- multilingual XLM-RoBERTa fine-tuned for code-mixed texts identification;
    \item ChatGPT-3.5 -- zero-shot setting for detecting code-mixed texts with use of OpenAI's ChatGPT-3.5;
    \item * ELMICT -- the model based on Ensemble Learning for Multilingual Identification of Code-mixed Texts.
\end{itemize}

We utilize three metrics to assess the effectiveness of classification models for detection of code-mixing, which are Accuracy (ACC), Area Under the Receiver Operating Characteristic Curve (AUC), and macro F-measure (F1), in alignment with practices of evaluation of binary text classification.

\section{Result Analysis and Discussion}

We begin by presenting the performance results of the proposed \textit{ELMICT} model compared to baseline schemes for research question RQ1. This is followed by an analysis of the proposed model's performance for cross-lingual classification for RQ2, and a detailed examination of social media content featuring code-mixing for RQ3.

\begin{table}
\scriptsize
\begin{tabularx}{\columnwidth}{|l|X|X|X|}
 \hline
 Model Scheme & ACC & F1 & AUC\\
 \hline
 \textit{lingua} & 78.70$\pm$2.18 & 62.39$\pm$3.16 & 73.20$\pm$5.28\\  
 \textit{Random Forest} & 71.74$\pm$0.08 & 71.55$\pm$0.27 & 71.85$\pm$0.64 \\
 \textit{Adaptive Boosting} & 69.57$\pm$5.75 & 69.17$\pm$6.21 & 69.77$\pm$6.55 \\
 \textit{Gradient Boosting} & 68.70$\pm$6.81 & 68.04$\pm$7.65 & 68.86$\pm$6.33 \\
 \textit{XLM-RoBERTa} & 97.05$\pm$0.62 & 92.08$\pm$1.74 & 91.60$\pm$1.23\\
 \textit{ChatGPT-3.5} & 77.45$\pm$1.62 & 54.82$\pm$3.28 & 67.28$\pm$4.26 \\
 \textit{* ELMICT} & \textbf{97.55}$\pm$\textbf{0.13} & \textbf{92.84}$\pm$\textbf{4.22} & \textbf{94.90}$\pm$\textbf{2.01}\\
 \hline
\end{tabularx}
\caption{Comparison with baselines. Results of binary classification. The best performances are in bold. Training and developing data is code-mixing messages in English-Finnish. 5-fold CV. $*$ denotes the proposed models.}
\label{table:fi-dev}
\end{table}

\subsection{English-Finnish Code-Mixing Detection}

Initially, we assess our model's performance on English-Finnish code-mixed messages.  We employ 5-fold cross-validation to randomly split the data into train-dev chunks in a 90-10 proportion. Table~\ref{table:fi-dev} illustrates the performance evaluation of the \textit{ELMICT} model compared to other schemes, addressing RQ1. The results indicate that our proposed model consistently outperforms the baselines across all metrics. Particularly noteworthy is the superior performance of the \textit{ELMICT} model compared to the fine-tuned \textit{XLM-RoBERTa} model. This underscores the significance of leveraging features generated by multiple tokenizers and a language detector module for code-mixing text detection tasks. Additionally, the performance of \textit{Random Forest} and \textit{lingua} models demonstrates that utilizing features without soft labels generated by fine-tuned pre-trained language models underperform. Moreover, the classification results highlight the limitations of \textit{ChatGPT-3.5} in zero-shot learning settings. While we don't explore fine-tuning or prompt-engineering approaches, it's plausible that further enhancements could improve the performance of LLMs. Lastly, the experiment reveals that \textit{Random Forest} outperforms other ensemble models like \textit{Adaptive Boosting} and \textit{Gradient Boosting}. We exclusively utilize \textit{Random Forest} for ensemble modeling in further experiments.

In addition to cross-validation, we test the higher performing models on a test batch, which contains data from different threads and is excluded from train and development batches used for models' training and fine-tuning. Additional experiments demonstrate the robustness of our model. The test batch includes 297 texts (131 code-mixed and 166 non-code-mixed texts). Table~\ref{table:fi-test} demonstrates comparable performance for the majority of schemes, and significant improvement in the performance of \textit{lingua} detector. 
Furthermore, the experiment on test data batch helps to prove the usage of \textit{ELMICT} for the classification English-Finnish dataset to answer RQ3. While we expected a drop in model performance because of possible overfitting, the performance is even higher.

\begin{table}
\scriptsize

\begin{tabularx}{\columnwidth}{|l|X|X|X|}
 \hline
 Model Scheme & ACC & F1 & AUC\\
 \hline
 \textit{lingua} & 81.82 & 81.64 & 81.55 \\ 
 \textit{XLM-RoBERTa} & 93.94 & 93.93 & 93.93 \\
 \textit{ChatGPT-3.5} & 74.75 & 74.58 & 74.54 \\
 \textit{* ELMICT} & \textbf{97.64} & \textbf{97.64} & \textbf{97.62}\\
 \hline
\end{tabularx}
\caption{Comparison with baselines. Results of binary classification. The best performances are in bold. Testing data is code-mixing messages in English-Finnish. $*$ denotes the proposed models.}
\label{table:fi-test}
\end{table}

%After analyzing the errors made by ELMICT, we discovered that the messages most commonly classified as false negatives were those that:

%Similarly, the types of messages most commonly classified as false positive were those that:

\subsection{Cross-lingual Code-Mixing Detection}

In addition to monolingual classification tasks, there are also cross-lingual classification settings where the languages in the training and testing data are different. To assess the proposed framework's cross-lingual capability, we utilize a zero-shot setting, where we train and validate classification schemes on the data of English-Finnish dataset and test the model  on the data from the other dataset (English-Korean or English-Spanish). For test data classification, we use the same models from the previous experiment. The complete findings of the cross-lingual classification are outlined in Table~\ref{table:cross-lingual}.

\begin{table}
\scriptsize

\begin{tabularx}{\columnwidth}{|l|X|X|X|}
 \hline
 Model Scheme & ACC & F1 & AUC\\
 \hline
 \multicolumn{4}{|l|}{English-Korean}\\
 \hline
 \textit{lingua} & 97.10 & 49.26 & 50.00 \\ 
 \textit{XLM-RoBERTa} & 93.00 & 63.84 & \textbf{74.65} \\
 \textit{ChatGPT-3.5} & 92.40 & 57.58 & 63.14 \\
 \textit{* ELMICT} & \textbf{96.70} & \textbf{66.80} & 64.85 \\
 \hline
 \multicolumn{4}{|l|}{English-Spanish}\\
 \hline
 \textit{lingua} & 87.80 & 46.75 & 50.00 \\ 
 \textit{XLM-RoBERTa} & 88.60 & \textbf{76.88} & \textbf{81.16} \\
 \textit{ChatGPT-3.5} & 48.80 & 44.38 & 64.14 \\
 \textit{* ELMICT} & \textbf{90.40} & 73.57 & 70.18 \\
 \hline
\end{tabularx}
\caption{Comparison with baselines. Results of cross-lingual binary classification. The best performances are in bold. Testing data is code-mixing messages in English-Korean and English-Spanish. $*$ denotes the proposed models.}
\label{table:cross-lingual}
\end{table}

In comparison to the other schemes, \textit{ELMICT} exhibits comparable performance with the fine-tuned XLM-RoBERTa model. \textit{ELMICT} demonstrates higher ACC for both datasets, while because of strong imbalance in both datasets, the other two metrics are more relevant. While for English-Korean messages \textit{ELMICT} has higher F1, for English-Spanish fine-tuned XLM-RoBERTa has higher F1. Moreover, XLM-RoBERTa demonstrates higher AUC for both datasets. While at the same time, two other schemes demonstrate random results with AUC equals 50\%.

\subsection{Topic Analysis}

\begin{figure}[t]
\centering
\includegraphics[width=\columnwidth]{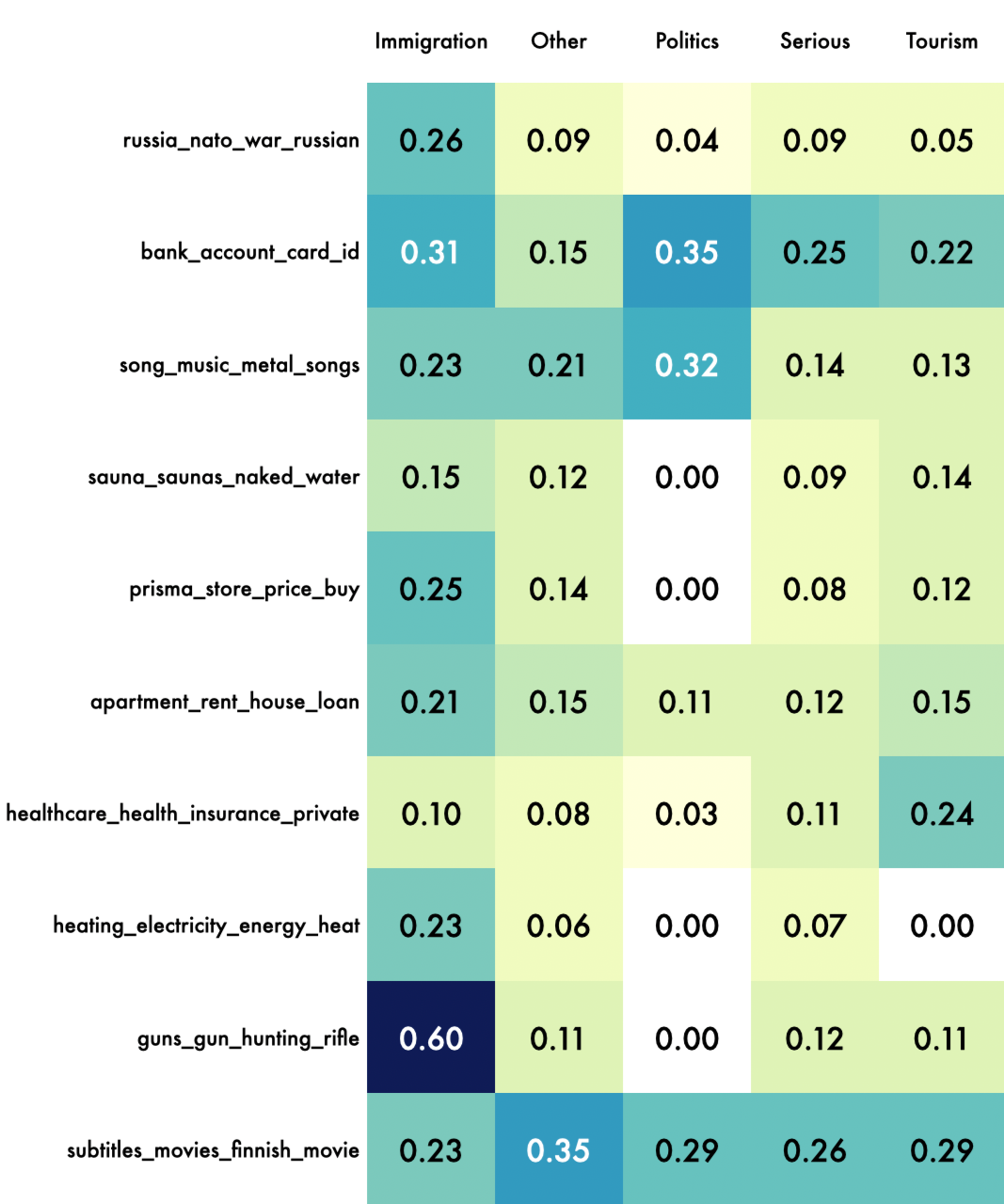}
\caption{Proportion of code-mixing messages per topic per flair in English-Finnish dataset.} 
\label{fig:topics_index}
\end{figure}

Figure~\ref{fig:topics_index} presents the top-10 most popular topics with use of code-mixing in threads with ``Immigration'' flair. The highest level of code-mixing usage in migration-related posts was detected in the topic related to guns (patruunatehdas -- cartridge factory; tarkkuuskivääri -- sniper rifle). The second most topic with high code-mixing usage is about employment and bank accounts in Finland (työ- ja elinkeinotoimisto (TE) -- Employment and Economic Development Office; pankki -- bank) because they are widely used not only in relation to financial services, but also for digital authentication in various services. The third topic with a high level of code-mixed messages is about the Russo-Ukrainian war and Finland's membership in NATO (siviilipalvelus -- civil service; taisteli puolella -- fought on the side). The other seven topics could be divided into two groups. The first one includes everyday life questions that could be addressed during migration: sauna (löyly -- steam), shopping (kierrätyskeskus -- recycling center), apartment renting (asunto -- apartment), healthcare (hoito -- care, therapy), and public utilities (pörssisähkö -- exchange electricity, also known as spot electricity). 

The second group is about popular cultural media content: local music and movie subtitles. These topics include many words and phrases in Finnish related to song and movie titles and artists' names. The latter group should not be classified as code-mixing because all these words and phrases are proper names. However, to avoid the classification of these messages as code-mixing is a separate challenging NLP-task. It could be tackled by applying a multilingual named entity recognition model in the future. This would have required additional experiments, though, which were beyond the scope of this paper.

\section{Conclusions and Future Work}

This paper explores code-mixing patterns in migration-related online conversations. Our proposed \textit{Ensemble Learning for Multilingual Identification of Code-mixed Text (ELMICT)} method allows for detection of messages with code-mixing in predominantly English-based datasets. 
The core idea of \textit{ELMICT} is its use in combination with multiple tokenizers outputs and soft labels generated by a pre-trained language model. 
The utilization of context-based soft labels allows us to predict code-mixing usage in migration-related contexts (everyday life challenges and cultural nuances), while tokenizer's outputs made models more robust in the new linguistic context.
Experiments on multiple English-based datasets that included code-mixing with words from Finnish, Korean, or Spanish show that the proposed model outperforms several baselines in the classification task.
Utilization of \textit{ELMICT} allows us to analyse the usage of code-mixing in migration-related threads on \textit{r/Finland} subreddit. 
The results of our analysis highlights a list of topics where code-mixing is highly predictable (housing market, shopping, public utilities, and healthcare), while also bringing to light particular topics (guns and hunting) and temporal discourses (Russo-Ukrainian war and NATO membership of Finland) where code-mixing was seen be be more widely used.

The \textit{ELMICT} model holds promising potential for application in public services that can utilize code-mixing in conversational agents and enhance trusting relations with migrants by appropriate usage of specific vocabularies.
This proposed model could be a part of the pipeline of model training in the Retrieval-Augmented Generation (RAG) module or database refinement.
By harnessing the capabilities of the \textit{ELMICT} model, organizations can strengthen their customer relationships by building trust based on communication and potentially innovate new solutions, grounded in the vocabulary that unites locals and migrants.
The versatility and adaptability of \textit{ELMICT} with the use of different language-related tokenizers beyond its initial scope could be one of the exploratory directions for future work.

There are certain limitations to our study that future work could address. 
First, the dataset we used for experimentation only contains messages posted during a limited time and contains information from only 3 subreddits. To improve the model's performance across various language pairs of code-mixing, it would be valuable to extend this dataset to include data collected over a longer period, more diverse topics, and additional languages.
Second, the topic analysis highlights the necessity of applying additional preprocessing steps, such as named entities recognition for proper names related to popular culture (titles, artists, etc.)
Third, our proposed model only identifies text contained in code-mixing, while for building conversational agents or any other application of code-mixed vocabulary, it's necessary to extract these tokens. Usage of \textit{ELMICT} will help to increase the efficiency of data annotation for the token classification task because of the automated filtering of monolingual texts.

\subsection{Reproducibility}
Datasets and code for the experiments described in this paper will be available for research purposes at the public repository \url{https://github.com/vitiugin/elmict}.

\section{Broader Impact and Ethics Statement}
For multilingual speakers, code-mixing is a communication method typically used when they are in a relaxed state and with people with whom they share close relationships. This linguistic strategy is employed specifically when the multilingual speaker has a trustful and intimate relationship with another person who shares similar linguistic and cultural backgrounds~\citep{wulandari2021code, kusumawati2023conversation}. In this context, to effectively build a trusting relationship between multilingual migrants, counselors, and conversational agents, incorporating the feature of code-mixing into the system is necessary. This adaptation would help migrants perceive that public services using human and conversational agents share similar linguistic and cultural backgrounds, fostering a sense of trust. However, we must ensure that such perceptions of trust induced by conversational agents using code-mixing in conversations do not make users believe such systems to be infallible or anthropomorphised; hence designing such systems to incorporate explainable outputs and accurate content is crucial.  

Furthermore, previous research has shown that multilingual users experience similar feelings of pressure when conversing with monolingual conversational agents as when conversing with strangers~\cite{choi2023toward}. This has led to the recognition that code-mixing conversational agents can provide multilingual users with a feeling of inclusion and acceptance in society. In recognizing the deep connection between social integration and trust formation during the migration process~\cite{dinesen2010rome}, it becomes evident that there is a significant opportunity for conversational agents to aid in this process. By providing a window of opportunity for migrants to build trust with public services, conversational agents can play a role in supporting migrants (and their human counselors) as they adapt to and integrate into their new environment. Aligning with this, it is of utmost importance to build relationships between migrants, human counselors, and conversational agents as part of a system of public services that can promote social integration and acceptance while migrants adapt to their new environment. As a result, these code-mixed digital offerings can be leveraged well to support the social integration of migrants, providing a pivotal step in promoting more inclusive public sector services.

We recognize that there are many ethical implications of this work related to discrimination, misuse and privacy of end-users. Furthermore, we assume that language identity of users such as code-mixing level could be used for profiling, may result in discriminatory practices. Since we demonstrate the potential to identify such attributes on social media, we are aware of how our research could be misused and abused, discriminating migrants~\cite{yim2019you, faingold2022language, ekwere2022language}.

To protect privacy, we refrain from disclosing sensitive personal information. Given Reddit's anonymous nature and the absence of mandatory personal data sharing, we commit to not sharing collected data that could be used to identify individuals. For reproducibility, we only provide comment IDs and code-mixing binary labels, keeping users' right to delete their data in the future if they choose.

We also strongly encourage future studies to consider the ethical dimensions of detecting language-related characteristics in social media texts, from study inception to final research dissemination.

\section*{Acknowledgments}
This work is supported by the Trust-M research project, a partnership between Aalto University, University of Helsinki, Tampere University, and the City of Espoo, funded in-part by a grant from the Strategic Research Council (SRC) in Finland. The authors also express their deep gratitude to CRAI-CIS research group members at Aalto University who helped in experimental setup and provided valuable insights.

% Use \bibliography{yourbibfile} instead or the References section will not appear in your paper
\bibliography{paper.bib}

\end{document}